\documentclass[letterpaper]{article}
\usepackage{bisidpreprint}
\usepackage[hyphens]{url}
\usepackage{graphicx}
\usepackage{natbib}
\usepackage{caption}
\usepackage{amsmath}
\usepackage{amssymb}
\usepackage{algorithm}
\usepackage{algorithmic}
\usepackage{newfloat}
\usepackage{listings}
\usepackage{booktabs}
\usepackage{array}
\usepackage{multirow}
\DeclareCaptionStyle{ruled}{labelfont=normalfont,labelsep=colon,strut=off}
\floatstyle{ruled}
\newfloat{listing}{tb}{lst}{}
\floatname{listing}{Listing}

\title{Which One Is Sounding? Binaural Sounding Instance Disambiguation for Audio-Visual Segmentation}
\author{
Haoze Ma\textsuperscript{1}, Yidi Li\textsuperscript{1,2}\corresponding, Yu Ma\textsuperscript{1}, Pengyuan Zhao\textsuperscript{1},\\
Yi Shi\textsuperscript{1}, Bin Ren\textsuperscript{3}, Yanwen Guo\textsuperscript{1}, Dimitrios Kanoulas\textsuperscript{2}
}
\affiliations{
\textsuperscript{1}Taiyuan University of Technology, China\\
\textsuperscript{2}University College London, United Kingdom\\
\textsuperscript{3}EPFL, Switzerland
}

\begin{document}

\maketitle

\begin{abstract}
Existing audio-visual segmentation methods mainly predict sound-related regions or category masks and often fail to identify the sounding instance when multiple same-class candidates coexist. Recent audio-visual instance segmentation methods address instance-level prediction but rely largely on motion and cross-frame association, which may become unreliable under sparse observations or inconspicuous motion. To this end, we introduce binaural audio, whose interaural differences encode source-direction information, to provide spatial constraints for distinguishing the sounding instance among same-class visual candidates. We propose Binaural Sounding Instance Disambiguation for Audio-Visual Segmentation (BiSID-AVS), which progressively refines category-level candidate responses into a mask of the true sounding instance. BiSID-AVS first grounds category-consistent candidate regions through audio prompting, then projects binaural cues into an image-plane spatial response map, and finally refines the true sounding instance through dual-level spatial conditioning at the feature and decoder levels. Experiments on the binaurally extended AVS-Synthetic benchmark and Habitat-AVS dataset for robotic indoor perception show that BiSID-AVS achieves the best performance on the multi-candidate subsets, improving instance-level mIoU by 35.19 and 41.21 percentage points, respectively, while substantially reducing false activation on non-sounding same-class regions.
\end{abstract}

\section{Introduction}

Audio-visual segmentation (AVS) aims to segment visual regions corresponding to synchronized audio. Since AVSBench, existing methods have addressed single-source, multi-source, and semantic segmentation settings~\cite{zhou2022avs,zhou2023avssem,yang2024combo,huang2025vct,gong2025avsmamba}. However, their predictions generally remain sound-related regions or category masks without explicit instance identity. This limitation becomes critical when multiple same-class candidates coexist. For example, audio semantics may indicate that a phone or alarm clock is sounding but cannot determine which particular instance produces the sound, causing a model to activate several same-class objects simultaneously. Recent robustness studies also suggest that some AVS models may rely heavily on visual saliency without sufficiently using audio context to identify the actual source~\cite{li2025robustavs}. Such category-level predictions are inadequate for robot interaction, source approaching, and object grasping, which require localization and segmentation of the specific sounding instance.

Audio-visual instance segmentation (AVIS) further identifies, segments, and tracks individual sounding objects~\cite{guo2025avism,zhu2026seavis}. Both AVIS and this work address the question of \textit{which instance is sounding}. Existing AVIS methods, however, mainly derive instance-discriminative evidence from continuous videos through frame-wise source localization, long-range audio-visual dependencies, and cross-frame instance association. These temporal cues may become unavailable or unreliable when only a single visual observation or sparse observations are provided, or when the sounding object exhibits little visible motion.

\begin{figure}[!t]
    \centering
    \includegraphics[width=\columnwidth]{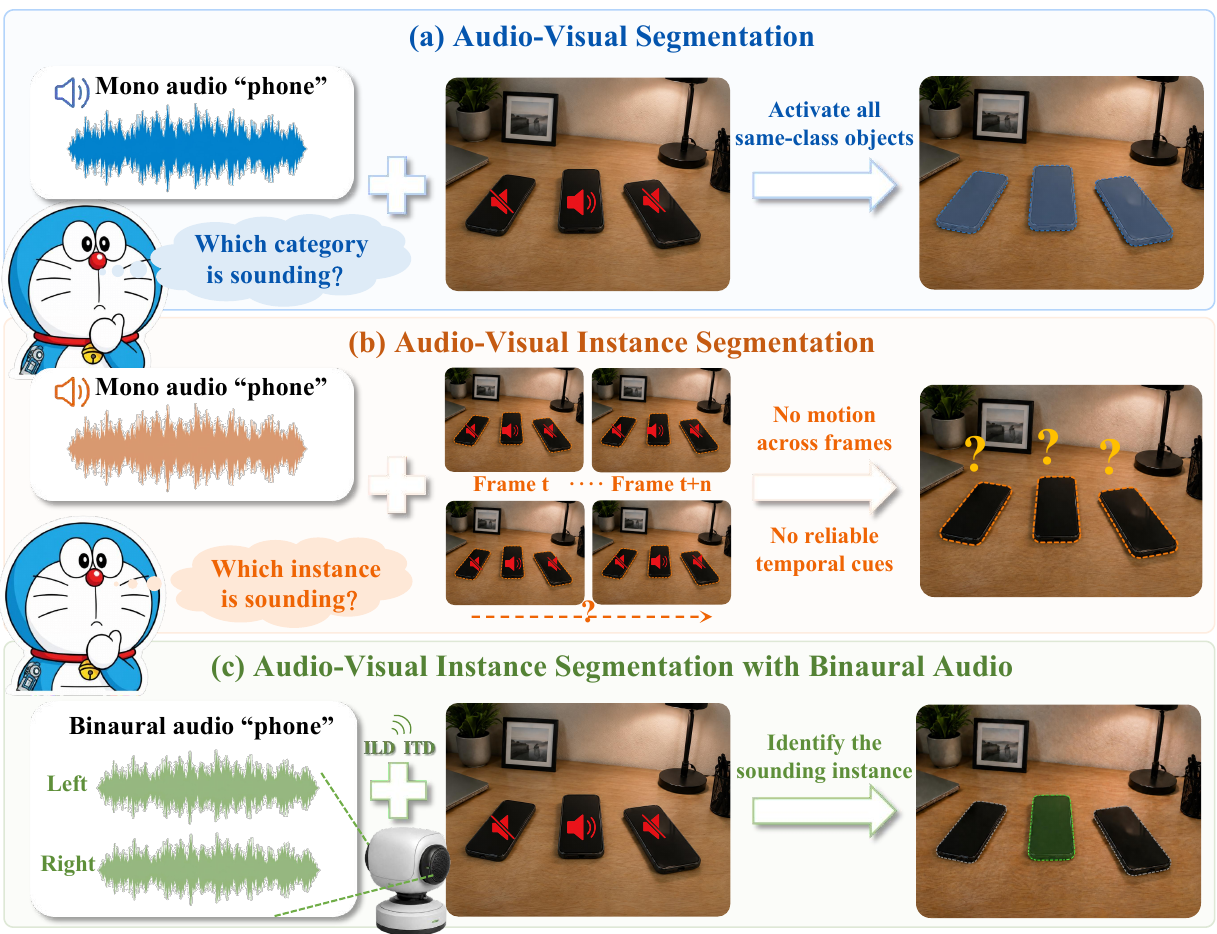}
    \caption{Task motivation of BiSID-AVS. AVS may activate multiple same-class candidates, while existing AVIS mainly relies on temporal context. BiSID-AVS instead uses binaural spatial cues to identify the true sounding instance.}
    \label{fig:motivation}
\end{figure}

\begin{figure*}[!t]
\centering
\includegraphics[width=\textwidth]{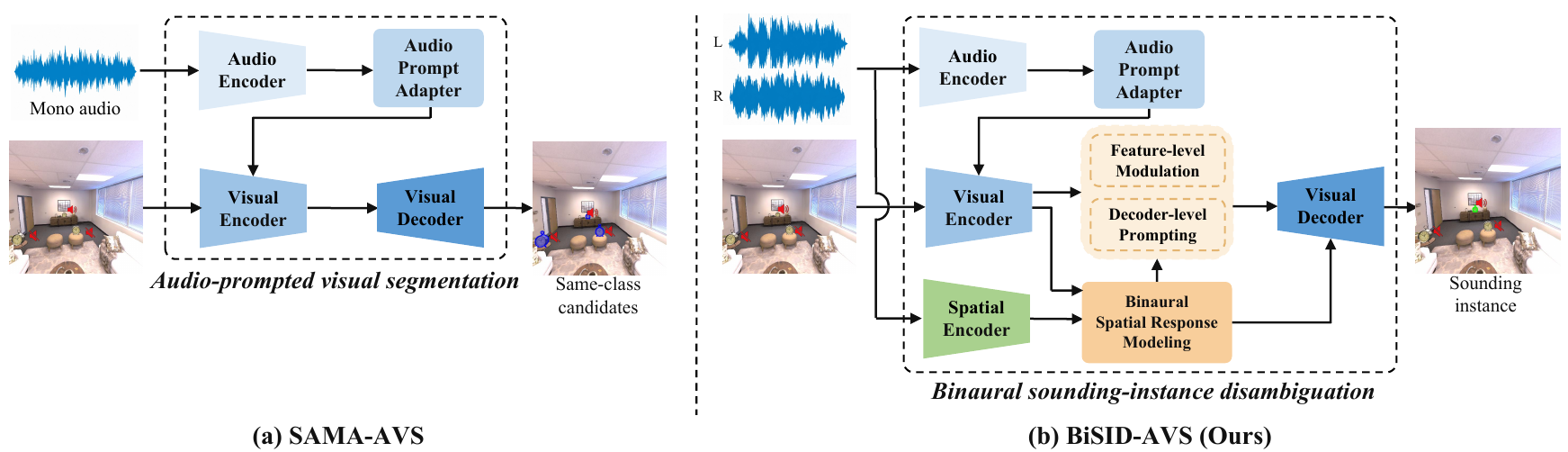}
\caption{Overall comparison between SAMA-AVS~\cite{liu2024sama} and BiSID-AVS. SAMA-AVS uses monaural audio prompting to model category-level audio-visual correspondence and may activate same-class candidates. BiSID-AVS models binaural spatial evidence and progressively refines category-consistent candidates into the mask of the true sounding instance.}
\label{fig:baseline_comparison}
\end{figure*}

Binaural audio provides spatial evidence complementary to video temporal information. Interaural time and level differences, together with direction-dependent spectral cues, encode the source location relative to the observer~\cite{grothe2010mechanisms}. Recent work further exploits the semantic and spatial information in binaural audio for visual sounding-object localization~\cite{li2024cyclic}, suggesting its potential as an instance-discriminative cue. Therefore, this work asks: \textit{Without reliable temporal cues, can a single visual observation and synchronized binaural audio identify the true sounding instance among multiple same-class candidates?} Figure~\ref{fig:motivation} illustrates the task motivation. AVS predicts sound-related regions or category masks, video-based AVIS mainly exploits temporal context to preserve instance identity, whereas our setting uses binaural spatial cues to disambiguate same-class candidates.

A straightforward solution is to independently estimate a source direction or spatial response map, generate visual proposals, and select the proposal with the highest spatial consistency. However, such decision-level fusion introduces spatial evidence only after mask generation. It therefore cannot guide candidate representation or mask formation and may remain unreliable when spatial responses are coarse, same-class objects are adjacent, or a visual proposal contains multiple instances.

To address this problem, we propose BiSID-AVS.
Existing audio-prompting methods such as SAMA-AVS ~\cite{liu2024sama} establish category-level audio-visual correspondence but lack explicit spatial evidence for distinguishing same-class instances. BiSID-AVS instead jointly models category-consistent candidates and binaural spatial evidence within a progressive instance-disambiguation process. It first grounds visual candidates associated with the sound category, then projects interaural cues into an image-plane spatial response, and finally incorporates this response into candidate refinement and mask prediction. This progressive process enables spatial evidence to participate directly in sounding-instance mask formation, rather than selecting among independently generated visual masks through post-hoc matching. Figure~\ref{fig:baseline_comparison} illustrates the transition from semantic audio-visual correspondence to binaural sounding-instance disambiguation.

Our main contributions are summarized as follows:
\begin{itemize}

\item To our knowledge, this is the first work to introduce binaural spatial cues into direct pixel-level audio-visual sounding-instance segmentation, enabling identification of the true sounding instance among same-class candidates when temporal cues are unavailable or unreliable.

\item We propose BiSID-AVS, a progressive framework that integrates candidate grounding, binaural spatial response modeling, and sounding-instance refinement. It directly incorporates spatial evidence into mask formation instead of relying on post-hoc matching between source localization and visual segmentation.

\item We construct Binaural AVS-Synthetic and Habitat-AVS for broad-category benchmarking and geometry-consistent robotic-view disambiguation. BiSID-AVS improves $\mathrm{mIoU}_{\mathrm{M}}$ by 35.19 and 41.21 percentage points on their multi-candidate subsets, respectively, while substantially reducing same-class false activation.

\end{itemize}

\begin{figure*}[t]
\centering
\includegraphics[width=\textwidth]{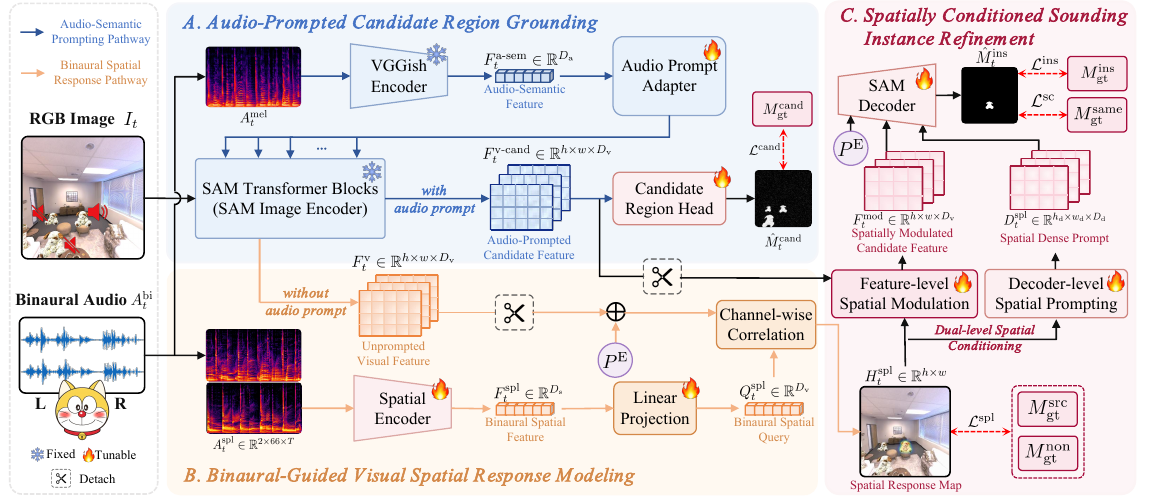}
\caption{Overall framework of BiSID-AVS. The model progressively performs (A) category-consistent candidate grounding, (B) binaural-guided image-plane spatial response modeling, and (C) sounding-instance mask refinement. Spatial evidence conditions both candidate-feature formation and mask decoding to refine category-consistent regions into the true sounding-instance mask.}
\label{fig:framework}
\end{figure*}

\section{Related work}

\paragraph{Audio-visual segmentation.} Audio-visual segmentation localizes sound-related image regions from synchronized audio. AVSBench established pixel-level protocols for single-source, multi-source, and semantic settings~\cite{zhou2022avs,zhou2023avssem}. Subsequent work advances cross-modal interaction through Transformer-based designs, query mechanisms, and multi-order bilateral relations~\cite{ling2023transavs,gao2023avsegformer,yang2024combo}; strengthens visual representations using vision-centric or state-space designs and foundation-model adaptation~\cite{huang2025vct,kirillov2023sam,liu2024sama,gong2025avsmamba}; and extends AVS to open-vocabulary and weakly supervised settings~\cite{guo2024ovavss,mo2023wsavs}. Among SAM-based approaches, SAMA-AVS uses audio prompting to ground category-consistent visual regions~\cite{liu2024sama}. However, monaural semantic prompting provides no explicit spatial evidence for distinguishing instances of the same class. 
Overall, existing AVS methods improve audio-visual correspondence but primarily produce sound-related region or category masks without explicit instance identity. They therefore do not directly determine which candidate is sounding when multiple same-class objects coexist in complex visual scenes.

\paragraph{Video-based audio-visual instance segmentation.} Audio-visual instance segmentation further requires identifying, segmenting, and tracking specific sounding targets. AVISM combines frame-wise source localization, instance-context extraction, and video-level matching to preserve instance identity, while SeaVIS studies online association and causal prediction in continuous video streams~\cite{guo2025avism,zhu2026seavis}. Existing AVIS methods therefore infer instance identity mainly from temporal context, frame-wise source evidence, and cross-frame instance association. In contrast, this work investigates whether binaural spatial cues can support same-class instance disambiguation when only a single or sparse visual observation is available.

\paragraph{Binaural spatial audio for embodied audio-visual perception.}
Binaural audio encodes source location through interaural time and level differences and direction-dependent spectral cues~\cite{grothe2010mechanisms}. Methods exploit audio-visual correspondence and binaural cues for spatial representation learning and sounding-object localization~\cite{gao2019visual,morgado2020spatial,chen2023sound,li2024cyclic}. SoundSpaces and SoundSpaces 2.0 support binaural rendering for embodied navigation, source localization, and acoustic reasoning~\cite{chen2020soundspaces,chen2022soundspaces2}. These methods predict source directions, navigation actions, or three-dimensional scene representations rather than image-plane sounding-instance masks. 
Closest to our setting, 3D audio-visual segmentation lifts 2D AVS masks into a 3D Gaussian representation and refines spatial clusters with binaural audio~\cite{sokolov2024threedavs}. BiSID-AVS instead incorporates binaural spatial evidence directly into two-dimensional mask formation, avoiding 2D-to-3D lifting and subsequent spatial-cluster refinement.

\section{Method}

BiSID-AVS formulates sounding-instance segmentation as a progressive same-class disambiguation process. Given a visual observation $I_t\in\mathbb{R}^{3\times H\times W}$ and synchronized binaural audio $A_t^{\mathrm{bi}}=(x_t^{\mathrm{L}},x_t^{\mathrm{R}})$, the model predicts a sounding-instance mask $\hat{M}_t^{\mathrm{ins}}\in[0,1]^{H\times W}$ that covers the true sounding instance among same-class candidates. As shown in Figure~\ref{fig:framework}, it first uses audio semantics to ground category-consistent candidate regions, then maps binaural spatial cues into a dense image-plane response, and finally incorporates this response into candidate refinement and mask decoding. The candidate-grounding and spatial-response pathways are specialized but jointly contribute to sounding-instance mask formation, avoiding post-hoc matching between independently estimated source locations and visual masks.

\subsection{Audio-Prompted Candidate Region Grounding}

This stage produces category-consistent candidate features using the SAM ViT-H image encoder~\cite{kirillov2023sam} and a frozen VGGish encoder~\cite{hershey2017cnn}. We average the binaural channels and extract an audio-semantic feature from the resulting mel spectrogram:
\begin{equation}
x_t^{\mathrm{mono}}=\frac{x_t^{\mathrm{L}}+x_t^{\mathrm{R}}}{2},\qquad
F_t^{\mathrm{a\text{-}sem}}
=E_{\mathrm{vgg}}\!\left(\mathcal{M}(x_t^{\mathrm{mono}})\right).
\end{equation}
At SAM layer $l$, a lightweight mapper $g_l$ converts the audio-semantic feature into a layer-wise prompt
$p_{t,l}=g_l(F_t^{\mathrm{a\text{-}sem}})$. The prompted branch broadcasts and adds $p_{t,l}$ to the visual tokens before the Transformer block, whereas the unprompted branch processes the visual tokens directly:
\begin{equation}
V_{t,l}^{\mathrm{v\text{-}cand}}
=B_l\!\left(V_{t,l-1}^{\mathrm{v\text{-}cand}}+p_{t,l}\right),
\qquad
V_{t,l}^{\mathrm{v}}=B_l(V_{t,l-1}^{\mathrm{v}}),
\end{equation}
where $B_l$ denotes the $l$-th SAM Transformer block. After the final block, the prompted and unprompted token sequences are reshaped and projected by the SAM encoder neck to obtain
the audio-prompted candidate feature $F_t^{\mathrm{v\text{-}cand}}$ and the unprompted visual feature $F_t^{\mathrm{v}}$.
The former represents audio-conditioned candidate regions, while the latter preserves category-agnostic spatial structure for subsequent spatial matching. An auxiliary head predicts the candidate-region map
$\hat{M}_t^{\mathrm{cand}}$, used only for auxiliary training.

\subsection{Binaural-Guided Visual Spatial Response Modeling}

The second stage estimates the spatial consistency between the binaural source and each image location. We transform the left and right waveforms into cochleagrams that preserve interaural differences~\cite{slaney1998auditory}, and stack them as the spatial-audio input:
\begin{equation}
C_t^{\mathrm{L}}=\mathcal{C}(x_t^{\mathrm{L}}),\quad
C_t^{\mathrm{R}}=\mathcal{C}(x_t^{\mathrm{R}}),\quad
A_t^{\mathrm{spl}}
=\operatorname{Stack}(C_t^{\mathrm{L}},C_t^{\mathrm{R}}).
\end{equation}
A lightweight convolutional encoder with temporal downsampling and global pooling extracts a spatial feature, which is projected into the visual space as a binaural spatial query:
\begin{equation}
F_t^{\mathrm{spl}}=E_{\mathrm{spl}}(A_t^{\mathrm{spl}})
\in\mathbb{R}^{D_{\mathrm{s}}},
\qquad
Q_t^{\mathrm{spl}}=W_{\mathrm{spl}}F_t^{\mathrm{spl}}
\in\mathbb{R}^{D_{\mathrm{v}}}.
\end{equation}

The query encodes receiver-relative source direction rather than a discrete image coordinate. To align it with image locations without interference from category-biased activations, we use the unprompted visual feature and add the native SAM positional encoding as an explicit image-plane reference:
\begin{equation}
\widetilde{F}_t^{\mathrm{v}}
=\operatorname{sg}(F_t^{\mathrm{v}})+P^{\mathrm{E}},
\qquad
P^{\mathrm{E}}\in\mathbb{R}^{h\times w\times D_{\mathrm{v}}},
\end{equation}
where $\operatorname{sg}(\cdot)$ denotes stop-gradient, which prevents spatial-response supervision from updating the visual backbone. A location-wise channel inner product then produces the spatial response map over the image plane:
\begin{equation}
H_t^{\mathrm{spl}}(u,v)
=\sum\nolimits_{d=1}^{D_{\mathrm{v}}}
\widetilde{F}_{t,d}^{\mathrm{v}}(u,v)
Q_{t,d}^{\mathrm{spl}}.
\end{equation}
The resulting spatial response map $H_t^{\mathrm{spl}}\in\mathbb{R}^{h\times w}$ assigns higher responses to locations more consistent with the binaural direction. Retaining this dense response, rather than regressing a discrete coordinate, allows uncertain spatial evidence to condition subsequent candidate refinement. 

\subsection{Spatially Conditioned Sounding-Instance Refinement}

The final stage resolves same-class ambiguity by conditioning both candidate features and mask decoding on the spatial response map to recover the true sounding instance.

\paragraph{Feature-level spatial modulation.}
We first convert the spatial response map $H_t^{\mathrm{spl}}$ into a bounded spatial gate and resize it to the candidate-feature resolution:
\begin{equation}
G_t^{\mathrm{spl}}
=\operatorname{Resize}\!\left(\sigma(H_t^{\mathrm{spl}})\right),
\qquad
G_t^{\mathrm{spl}}\in\mathbb{R}^{h\times w}.
\end{equation}
The candidate feature is then modulated in a residual form:
\begin{equation}
F_t^{\mathrm{mod}}
=\operatorname{sg}\!\left(F_t^{\mathrm{v\text{-}cand}}\right)
\odot\left(1+\alpha G_t^{\mathrm{spl}}\right),
\end{equation}
where $G_t^{\mathrm{spl}}$ is broadcast along the channel dimension, and
$F_t^{\mathrm{mod}}$
denotes the spatially modulated candidate feature.
Stop-gradient preserves candidate grounding from instance-level objectives.
The learnable global scalar $\alpha$ is initialized to zero, making the modulation initially an identity mapping.

\paragraph{Decoder-level spatial prompting.}
Feature modulation reweights the candidate representation, while decoder-level prompting provides an explicit spatial condition for mask generation. A lightweight mapper $\psi$ converts the response map into a dense prompt:
\begin{equation}
D_t^{\mathrm{spl}}=\psi(H_t^{\mathrm{spl}}),
\qquad
D_t^{\mathrm{spl}}
\in\mathbb{R}^{h_{\mathrm{d}}\times w_{\mathrm{d}}\times D_{\mathrm{d}}},
\end{equation}
whose dimensions match the SAM dense-prompt interface. The SAM decoder $\mathcal{D}_{\mathrm{SAM}}$ jointly receives the modulated feature, positional encoding, and spatial prompt:
\begin{equation}
P_t^{\mathrm{ins}}
=\mathcal{D}_{\mathrm{SAM}}
\left(
F_t^{\mathrm{mod}},
P^{\mathrm{E}},
D_t^{\mathrm{spl}}
\right),
\end{equation}
where $P_t^{\mathrm{ins}}$ denotes the sounding-instance logits, and no sparse prompt is used in mask decoding.

\subsection{Training Objectives for Instance Disambiguation}

BiSID-AVS is trained end-to-end with four complementary objectives for sounding-instance prediction, candidate grounding, spatial-response learning, and same-class distractor suppression:
\begin{equation}
\mathcal{L}
= \lambda^{\mathrm{ins}}\mathcal{L}^{\mathrm{ins}}
+ \lambda^{\mathrm{cand}}\mathcal{L}^{\mathrm{cand}}
+ \lambda^{\mathrm{spl}}\mathcal{L}^{\mathrm{spl}}
+ \lambda^{\mathrm{sc}}\mathcal{L}^{\mathrm{sc}},
\end{equation}
where the $\lambda$ terms balance the four objectives. Given the sounding-instance logits $P_t^{\mathrm{ins}}$, candidate-region logits $P_t^{\mathrm{cand}}$, and spatial response map $H_t^{\mathrm{spl}}$, their probability maps are
$\hat{M}_t^{\mathrm{ins}}=\sigma(P_t^{\mathrm{ins}})$,
$\hat{M}_t^{\mathrm{cand}}=\sigma(P_t^{\mathrm{cand}})$, and
$\hat{S}_t^{\mathrm{spl}}=\sigma(H_t^{\mathrm{spl}})$.
The instance and candidate predictions are upsampled to the input resolution before computing the corresponding mask losses.

\paragraph{Mask losses.}
The sounding-instance and candidate-region predictions are supervised by
their corresponding masks:
\begin{equation}
\begin{aligned}
\mathcal{L}^{\mathrm{ins}}
={}&\mathcal{L}_{\mathrm{bce}}^{\mathrm{logit}}
(P_t^{\mathrm{ins}},M_{\mathrm{gt}}^{\mathrm{ins}})
+\mathcal{L}_{\mathrm{iou}}
(\hat{M}_t^{\mathrm{ins}},M_{\mathrm{gt}}^{\mathrm{ins}}),\\
\mathcal{L}^{\mathrm{cand}}
={}&\mathcal{L}_{\mathrm{bbce}}^{\mathrm{logit}}
(P_t^{\mathrm{cand}},M_{\mathrm{gt}}^{\mathrm{cand}})
+\mathcal{L}_{\mathrm{iou}}
(\hat{M}_t^{\mathrm{cand}},M_{\mathrm{gt}}^{\mathrm{cand}}).
\end{aligned}
\end{equation}
Here, $\mathcal{L}_{\mathrm{bbce}}^{\mathrm{logit}}$ denotes BCE on logits with the positive-class weight $\min(N_{-}/N_{+},100)$, where $N_{+}$ and $N_{-}$ are the foreground and background pixel counts, and $\mathcal{L}_{\mathrm{iou}}$ is one minus the soft intersection-over-union between the predicted probability map and ground-truth mask.

\paragraph{Spatial response map loss.}
We area-downsample the sounding instance mask and retain locations with positive target coverage as supervision on the spatial-response grid:
\begin{equation}
M_{\mathrm{gt}}^{\mathrm{src}}
=\mathbb{I}\!\left[
\operatorname{Down}_{\mathrm{area}}
(M_{\mathrm{gt}}^{\mathrm{ins}})>0
\right],
\qquad
M_{\mathrm{gt}}^{\mathrm{non}}
=1-M_{\mathrm{gt}}^{\mathrm{src}}.
\end{equation}
This prevents small targets from disappearing at the response-map
resolution. The spatial loss balances source enhancement and non-source
suppression:
\begin{equation}
\mathcal{L}^{\mathrm{spl}}
=
-\frac{
\left\langle M_{\mathrm{gt}}^{\mathrm{src}},
\log \hat{S}_t^{\mathrm{spl}}\right\rangle}
{|M_{\mathrm{gt}}^{\mathrm{src}}|+\epsilon}
-\beta^{\mathrm{non}}
\frac{
\left\langle M_{\mathrm{gt}}^{\mathrm{non}},
\log(1-\hat{S}_t^{\mathrm{spl}})\right\rangle}
{|M_{\mathrm{gt}}^{\mathrm{non}}|+\epsilon},
\end{equation}
where $\langle\cdot,\cdot\rangle$ denotes the sum of element-wise
products over the spatial-response grid. The first term increases
responses within the true source region, while the second suppresses
responses elsewhere. Separate normalization balances the sparse source
region and the dominant non-source region, with $\beta^{\mathrm{non}}$
controlling the suppression strength.

\paragraph{Same-class suppression loss.}
To explicitly penalize non-sounding same-class regions, the loss is defined as:
\begin{equation}
\begin{aligned}
M_{\mathrm{gt}}^{\mathrm{same}}
&=M_{\mathrm{gt}}^{\mathrm{cand}}
\odot(1-M_{\mathrm{gt}}^{\mathrm{ins}}),\\
\mathcal{L}^{\mathrm{sc}}
&=\frac{
\sum_{(u,v)\in\Omega_{\mathrm{img}}}
\hat{M}_t^{\mathrm{ins}}(u,v)
M_{\mathrm{gt}}^{\mathrm{same}}(u,v)}
{|M_{\mathrm{gt}}^{\mathrm{same}}|+\epsilon},
\end{aligned}
\end{equation}
where $\Omega_{\mathrm{img}}$ denotes the full-resolution image grid used for mask supervision.
$\mathcal{L}^{\mathrm{sc}}$ is set to zero when
$M_{\mathrm{gt}}^{\mathrm{same}}$ is empty.

All losses are jointly optimized. $\mathcal{L}^{\mathrm{cand}}$
supervises candidate grounding, $\mathcal{L}^{\mathrm{spl}}$ directly
supervises the spatial encoder and query projection, and
$\mathcal{L}^{\mathrm{ins}}$ and $\mathcal{L}^{\mathrm{sc}}$ further
update the spatial-response pathway, conditioning modules, and mask
decoder. Stop-gradient isolates the unprompted visual anchor and
candidate-grounding pathway from unintended updates.
\section{Experiments}

\subsection{Datasets and Experimental Setup}
\paragraph{Datasets.}
To evaluate sounding-instance disambiguation under controlled and geometry-consistent conditions, we use two complementary datasets. Both provide RGB images, synchronized binaural audio, sounding-instance masks and boxes, and same-class candidate masks. The sounding-instance annotation identifies the audio-associated target, and the candidate mask covers all instances of the sound category.

Binaural AVS-Synthetic extends AVS-Synthetic~\cite{liu2024sama}, a commonly used synthetic benchmark for audio-visual segmentation, with instance-level annotations and binaural audio. We verify instances in the original semantic masks using GroundingDINO~\cite{liu2024groundingdino}, select one as the sounding target, and derive its rendering parameters from image position and relative depth estimated by Depth Anything~\cite{yang2024depthanything}. Binaural audio is synthesized using the SADIE D1 HRTF~\cite{armstrong2018sadie}. The dataset contains 31,045 samples from 46 categories, including 14.5\% multi-candidate samples.

Habitat-AVS is constructed with Habitat, SoundSpaces, and Replica~\cite{savva2019habitat,chen2020soundspaces,chen2022soundspaces2,straub2019replica}. Habitat provides image-aligned instance annotations, while SoundSpaces renders binaural audio from the target's three-dimensional location. The dataset contains 11,781 samples from 14 categories across four indoor scenes and ten predefined robotic viewpoints, with 45.3\% containing multiple same-class candidates. Together, Binaural AVS-Synthetic provides broad-category evaluation under controlled image--audio spatial correspondence, whereas Habitat-AVS evaluates geometry-consistent disambiguation with more realistic three-dimensional acoustics and robotic viewpoints.

\paragraph{Evaluation metrics.}
We report instance-level mIoU, computed as the mean per-sample IoU,
mean Sounding Instance Localization Accuracy (mSILA) adapted from
Frame-level Sound Location Accuracy (FSLA)~\cite{guo2025avism}, same-class false activation
$F_{\mathrm{same}}$, and the Same-Class Disambiguation Score (SC-DS).
Given the binarized prediction
$\widetilde{M}_i^{\mathrm{ins}}
=\mathbb{I}[\hat{M}_i^{\mathrm{ins}}\geq0.5]$, the metrics are:
\begin{equation}
\begin{aligned}
\operatorname{mSILA}
&=\frac{1}{N|\mathcal{T}|}
\sum_{i=1}^{N}\sum_{\tau\in\mathcal{T}}
\mathbb{I}\!\left[
\operatorname{IoU}(\widetilde{M}_i^{\mathrm{ins}},
M_{\mathrm{gt},i}^{\mathrm{ins}})\geq\tau
\right],\\
F_{\mathrm{same}}^{(i)}
&=\frac{
|\widetilde{M}_i^{\mathrm{ins}}\cap
M_{\mathrm{gt},i}^{\mathrm{same}}|}
{|M_{\mathrm{gt},i}^{\mathrm{same}}|+\epsilon},\\
\operatorname{SC\mbox{-}DS}^{(i)}
&=\operatorname{IoU}(\widetilde{M}_i^{\mathrm{ins}},
M_{\mathrm{gt},i}^{\mathrm{ins}})
\left(1-F_{\mathrm{same}}^{(i)}\right),
\end{aligned}
\end{equation}
where $\mathcal{T}=\{0.50,0.55,\ldots,0.95\}$ and
$M_{\mathrm{gt},i}^{\mathrm{same}}
=M_{\mathrm{gt},i}^{\mathrm{cand}}
\odot(1-M_{\mathrm{gt},i}^{\mathrm{ins}})$
denotes the non-sounding same-class region.
$F_{\mathrm{same}}$ and SC-DS are averaged over multi-candidate samples. The subscripts $\mathrm{S}$, $\mathrm{M}$, and $\mathrm{All}$ denote the single-candidate, multi-candidate, and full sets, respectively.

\paragraph{Protocols and implementation details.}
We use five-fold cross-validation on both datasets. Binaural AVS-Synthetic is grouped by original sample and stratified by category and candidate setting. For Habitat-AVS, the ten viewpoints form five fixed pairs, with one pair held out per fold for viewpoint-disjoint evaluation; we also perform four-fold leave-one-scene-out evaluation. In each fold, 10\% of the training data is reserved for validation, and the checkpoint with the highest validation $\mathrm{mIoU}_{\mathrm{All}}$ is used for testing.
BiSID-AVS uses SAM ViT-H initialized from the official checkpoint and VGGish pretrained on AudioSet. The VGGish and SAM image encoders are frozen, while the SAM mask decoder and all newly introduced modules are optimized. Images are resized to $512\times512$, and 2-s audio clips sampled at 16 kHz are converted into cochleagrams using 66 filters spanning 50--8000 Hz.
We set $\lambda^{\mathrm{ins}}=\lambda^{\mathrm{cand}}=1.0$, $\lambda^{\mathrm{spl}}=0.5$, $\lambda^{\mathrm{sc}}=0.1$, and $\beta^{\mathrm{non}}=1.0$. Training uses AdamW for 40 epochs with an initial learning rate of $2\times10^{-4}$, weight decay of 0.01, cosine decay without warm-up, and batch size 1. Each fold is trained once with a fixed random seed on one NVIDIA RTX 4090 GPU. Details are provided in Appendix~\ref{sec:supp_implementation}.

\begin{table*}[t]
\centering
\footnotesize
\setlength{\tabcolsep}{1pt}
\renewcommand{\arraystretch}{1.05}
\begin{tabular*}{\textwidth}{@{\extracolsep{\fill}}clcccccccc@{}}
\toprule
Dataset & Method &
$\mathrm{mIoU}_{\mathrm{S}}^{\uparrow}$ &
$\mathrm{mIoU}_{\mathrm{M}}^{\uparrow}$ &
$\mathrm{mIoU}_{\mathrm{All}}^{\uparrow}$ &
$\mathrm{mSILA}_{\mathrm{S}}^{\uparrow}$ &
$\mathrm{mSILA}_{\mathrm{M}}^{\uparrow}$ &
$\mathrm{mSILA}_{\mathrm{All}}^{\uparrow}$ &
$F_{\mathrm{same},\mathrm{M}}^{\downarrow}$ &
$\mathrm{SC\mbox{-}DS}_{\mathrm{M}}^{\uparrow}$ \\
\midrule
\multirow{7}{*}{\shortstack{Binaural\\AVS-Synthetic}}
& TPAVI & 0.7838 & 0.2742 & 0.7099 & 0.7587 & 0.0985 & 0.6629 & 0.6841 & \underline{0.1184} \\
& AVSegFormer & \textbf{0.8619} & 0.3275 & 0.7844 & \textbf{0.8213} & \underline{0.1308} & \underline{0.7211} & 0.7160 & 0.1067 \\
& COMBO & 0.7695 & 0.2641 & 0.6962 & 0.7397 & 0.0953 & 0.6462 & \underline{0.6752} & 0.1037 \\
& VCT & \underline{0.8603} & \underline{0.3482} & \underline{0.7860} & \underline{0.8178} & 0.1297 & 0.7183 & 0.7227 & 0.0598 \\
& AVISM & 0.8217 & 0.3376 & 0.7515 & 0.7839 & 0.1035 & 0.6852 & 0.7065 & 0.0724 \\
& SAMA-AVS & 0.8426 & 0.3339 & 0.7688 & 0.8093 & 0.1187 & 0.7091 & 0.7143 & 0.0916 \\
& BiSID-AVS (Ours) & 0.8386 & \textbf{0.7001} & \textbf{0.8185} & 0.7942 & \textbf{0.6099} & \textbf{0.7674} & \textbf{0.1165} & \textbf{0.6544} \\
\midrule
\multirow{7}{*}{Habitat-AVS}
& TPAVI & 0.9257 & 0.4670 & 0.7177 & 0.8998 & 0.1250 & 0.5486 & 0.9583 & 0.0209 \\
& AVSegFormer & \textbf{0.9448} & \underline{0.4831} & \underline{0.7356} & \underline{0.9517} & \underline{0.1421} & 0.5849 & 0.9780 & 0.0112 \\
& COMBO & 0.8828 & 0.4473 & 0.6853 & 0.8197 & 0.1069 & 0.4968 & \underline{0.9278} & \underline{0.0341} \\
& VCT & \underline{0.9375} & 0.4714 & 0.7262 & 0.9222 & 0.1290 & 0.5625 & 0.9629 & 0.0185 \\
& AVISM & 0.9137 & 0.4528 & 0.7049 & 0.8725 & 0.1172 & 0.5303 & 0.9308 & 0.0293 \\
& SAMA-AVS & 0.9368 & 0.4811 & 0.7303 & \textbf{0.9544} & 0.1401 & \underline{0.5855} & 0.9762 & 0.0123 \\
& BiSID-AVS (Ours) & 0.9321 & \textbf{0.8952} & \textbf{0.9154} & 0.9263 & \textbf{0.8564} & \textbf{0.8946} & \textbf{0.0356} & \textbf{0.8811} \\
\midrule
\multirow{2}{*}{\shortstack{Habitat-AVS\\(scene-disjoint)}}
& SAMA-AVS & \textbf{0.9219} & \underline{0.4537} & \underline{0.7098} & \textbf{0.9129} & \underline{0.1042} & \underline{0.5465} & \underline{0.9413} & \underline{0.0249} \\
& BiSID-AVS (Ours) & \underline{0.9147} & \textbf{0.8543} & \textbf{0.8873} & \underline{0.8937} & \textbf{0.8216} & \textbf{0.8610} & \textbf{0.0437} & \textbf{0.8473} \\
\bottomrule
\end{tabular*}
\caption{Comparison of BiSID-AVS with representative audio-visual segmentation and audio-visual instance segmentation methods. The final block reports four-fold leave-one-scene-out evaluation on Habitat-AVS. S, M, and All denote the single-candidate subset, multi-same-class-candidate subset, and all samples, respectively.}
\label{tab:main_results}
\vspace{-1.5mm}
\end{table*}

\subsection{Comparison with Existing Methods}
\paragraph{Main quantitative comparison.} We compare BiSID-AVS with representative AVS methods, including
TPAVI~\cite{zhou2022avs}, AVSegFormer~\cite{gao2023avsegformer},
COMBO~\cite{yang2024combo}, VCT~\cite{huang2025vct}, and
SAMA-AVS~\cite{liu2024sama}, as well as the sounding-instance method
AVISM~\cite{guo2025avism}. All baselines are retrained on identical
splits using public implementations and recommended settings, with only
necessary input and output adaptations. Monaural methods use the
channel-averaged audio, while video-based methods receive repeated
static frames. Following its original protocol, AVISM additionally uses
the sound category and number of sounding instances.
As shown in Table~\ref{tab:main_results}, BiSID-AVS improves
$\mathrm{mIoU}_{\mathrm{M}}$ by 35.19 and 41.21 percentage points and
$\mathrm{mSILA}_{\mathrm{M}}$ by 47.91 and 71.43 points on two datasets. It also achieves the lowest
$F_{\mathrm{same},\mathrm{M}}$ and highest
$\mathrm{SC\mbox{-}DS}_{\mathrm{M}}$ on both datasets, demonstrating
stronger sounding-instance coverage and same-class distractor
suppression. On single-candidate samples, where candidate grounding is
usually sufficient, spatial conditioning brings limited gains and may
slightly reduce mask completeness.

\begin{table}[t]
\centering
\footnotesize
\setlength{\tabcolsep}{0pt}
\renewcommand{\arraystretch}{1.05}
\begin{tabular*}{\columnwidth}{@{\extracolsep{\fill}}lcccc@{}}
\toprule
Method &
$\mathrm{mIoU}_{\mathrm{M}}^{\uparrow}$ &
$\mathrm{mSILA}_{\mathrm{M}}^{\uparrow}$ &
$F_{\mathrm{same},\mathrm{M}}^{\downarrow}$ &
$\mathrm{SC\mbox{-}DS}_{\mathrm{M}}^{\uparrow}$ \\
\midrule
SSL-HM + VP & 0.2956 & 0.2383 & \underline{0.1429}& 0.2849 \\
Cand. + SSL-HM + VP & 0.7599 & 0.6731 & 0.1554& 0.7419\\
GT Cand. + SSL-HM
& \textbf{0.9326} & \textbf{0.8928} & 0.1695& \underline{0.8617}\\
BiSID-AVS (Ours) & \underline{0.8952} & \underline{0.8564} & \textbf{0.0356} & \textbf{0.8811} \\
\bottomrule
\end{tabular*}
\caption{Decision-level fusion baselines and oracle analysis on the Habitat-AVS multi-candidate subset. GT Cand. denotes the ground-truth candidate mask.}
\label{tab:decision_fusion}
\vspace{-2.2mm}
\end{table}

\paragraph{Scene-disjoint generalization.}
Under the leave-one-scene-out protocol, both methods show performance
degradation due to unseen visual layouts and acoustic environments.
BiSID-AVS substantially outperforms SAMA-AVS on the multi-candidate
and full sets and achieves markedly better same-class disambiguation,
while remaining slightly lower on the single-candidate metrics. These
results demonstrate stronger generalization to unseen scenes in the
challenging same-class disambiguation setting.
\subsection{Analysis of Binaural Spatial Disambiguation}

\paragraph{Comparison with decision-level fusion.}
We construct three output-stage baselines using the upsampled spatial
response as a Sound Source Localization Heatmap (SSL-HM) and Visual
Proposals (VP) from frozen SAM ViT-H. SSL-HM + VP selects the proposal
with the highest mean response. Cand. Map + SSL-HM + VP weights the
heatmap with the predicted candidate-region probability map, whereas GT
Cand. + SSL-HM restricts it using the ground-truth candidate mask.
All baselines incorporate spatial evidence only through post-hoc output-stage fusion.
As shown in Table~\ref{tab:decision_fusion}, candidate-map weighting
substantially improves $\mathrm{mIoU}_{\mathrm{M}}$ but increases
$F_{\mathrm{same},\mathrm{M}}$, indicating stronger activation of
same-class distractors. The GT-candidate baseline achieves the highest
$\mathrm{mIoU}_{\mathrm{M}}$ and $\mathrm{mSILA}_{\mathrm{M}}$, yet
retains an $F_{\mathrm{same},\mathrm{M}}$ of 0.1695. Without GT information, BiSID-AVS obtains the lowest
$F_{\mathrm{same},\mathrm{M}}$ and highest
$\mathrm{SC\mbox{-}DS}_{\mathrm{M}}$, showing that spatial conditioning
during refinement and decoding suppresses distractors more effectively
than post-hoc selection.

\begin{table}[t]
\centering
\footnotesize
\setlength{\tabcolsep}{0pt}
\renewcommand{\arraystretch}{1.05}
\begin{tabular*}{\columnwidth}{@{\extracolsep{\fill}}lcccc@{}}
\toprule
Audio condition &
$\mathrm{mIoU}_{\mathrm{M}}^{\uparrow}$ &
$\mathrm{mSILA}_{\mathrm{M}}^{\uparrow}$ &
$F_{\mathrm{same},\mathrm{M}}^{\downarrow}$ &
$\mathrm{SC\mbox{-}DS}_{\mathrm{M}}^{\uparrow}$ \\
\midrule
Original
& \textbf{0.8952} & \textbf{0.8564} & \textbf{0.0356} & \textbf{0.8811} \\
Swapped channels
& \underline{0.1737} & \underline{0.1297} & \underline{0.4483} & \underline{0.1685} \\
Mismatched audio
& 0.1141 & 0.0749 & 0.6437 & 0.1054 \\
\bottomrule
\end{tabular*}
\caption{Effect of interaural spatial cues on the Habitat-AVS multi-candidate subset.}
\label{tab:interaural_cues}
\vspace{-2.5mm}
\end{table}

\paragraph{Effect of interaural spatial cues.}
To verify that BiSID-AVS exploits binaural spatial evidence for same-class disambiguation, we perturb the audio input while keeping the image fixed. Channel swapping reverses the left--right structure, whereas mismatched audio uses a same-scene, same-category sample from another source location. As shown in Table~\ref{tab:interaural_cues}, both perturbations markedly reduce $\mathrm{mIoU}_{\mathrm{M}}$ and $\mathrm{mSILA}_{\mathrm{M}}$ and increase $F_{\mathrm{same},\mathrm{M}}$, with mismatched audio causing the largest degradation. This confirms that BiSID-AVS relies on interaural structure and audio--visual spatial correspondence, rather than sound-category semantics alone. 

\subsection{Ablation Studies}

\paragraph{Architecture ablation.}
Table~\ref{tab:ablation_structure} evaluates the two functional pathways and the two components of dual-level spatial conditioning. Cfg. A retains only Candidate Grounding (CG) and performs poorly on same-class disambiguation, showing that category-consistent candidates alone are insufficient. Cfg. B combines Spatial Response (SR) with Decoder-level Prompting (DP), since SR
does not directly produce a mask, and achieves strong localization
performance without candidate grounding. 
Cfg. C retains CG and
SR but disables both conditioning mechanisms. Its weaker results show
that computing candidate and spatial features alone does not effectively
integrate spatial evidence into mask prediction. 
Cfg. D and E
add Feature-level Modulation (FM) and DP separately. FM suppresses spatially inconsistent candidate features, whereas DP directly conditions mask decoding. 
Compared with B, E improves target coverage through candidate grounding but may also increase activation on same-class distractors. Cfg. F combines FM and DP and achieves the best overall performance, confirming that feature modulation and decoder prompting provide complementary ways to incorporate binaural spatial evidence.

\begin{table}[t]
\centering
\footnotesize
\setlength{\tabcolsep}{1pt}
\renewcommand{\arraystretch}{1.05}
\begin{tabular*}{\columnwidth}{@{\extracolsep{\fill}}ccccccccc@{}}
\toprule
Cfg. & CG & SR & FM & DP &
$\mathrm{mIoU}_{\mathrm{M}}^{\uparrow}$ &
$\mathrm{mSILA}_{\mathrm{M}}^{\uparrow}$ &
$F_{\mathrm{same},\mathrm{M}}^{\downarrow}$ &
$\mathrm{SC\mbox{-}DS}_{\mathrm{M}}^{\uparrow}$ \\
\midrule
A & $\checkmark$ & -- & -- & -- & 0.4799 & 0.1394 & 0.9762 & 0.0123 \\
B & -- & $\checkmark$ & -- & $\checkmark$ & 0.8427 & 0.8132 & \underline{0.0461} & 0.8365 \\
C & $\checkmark$ & $\checkmark$ & -- & -- & 0.8391& 0.8091& 0.1273& 0.8227\\
D & $\checkmark$ & $\checkmark$ & $\checkmark$ & -- & 0.8532& 0.8197& 0.0842& 0.8415\\
E & $\checkmark$ & $\checkmark$ & -- & $\checkmark$ & \underline{0.8676} & \underline{0.8379} & 0.1039 & \underline{0.8593} \\
F & $\checkmark$ & $\checkmark$ & $\checkmark$ & $\checkmark$ & \textbf{0.8952} & \textbf{0.8564} & \textbf{0.0356} & \textbf{0.8811} \\
\bottomrule
\end{tabular*}
\caption{Architecture ablation. CG: candidate grounding, SR: spatial response modeling, FM: feature-level modulation, DP: decoder-level prompting.}
\label{tab:ablation_structure}
\vspace{-1mm}
\end{table}

\begin{figure}[t]
\centering
\includegraphics[width=\columnwidth]{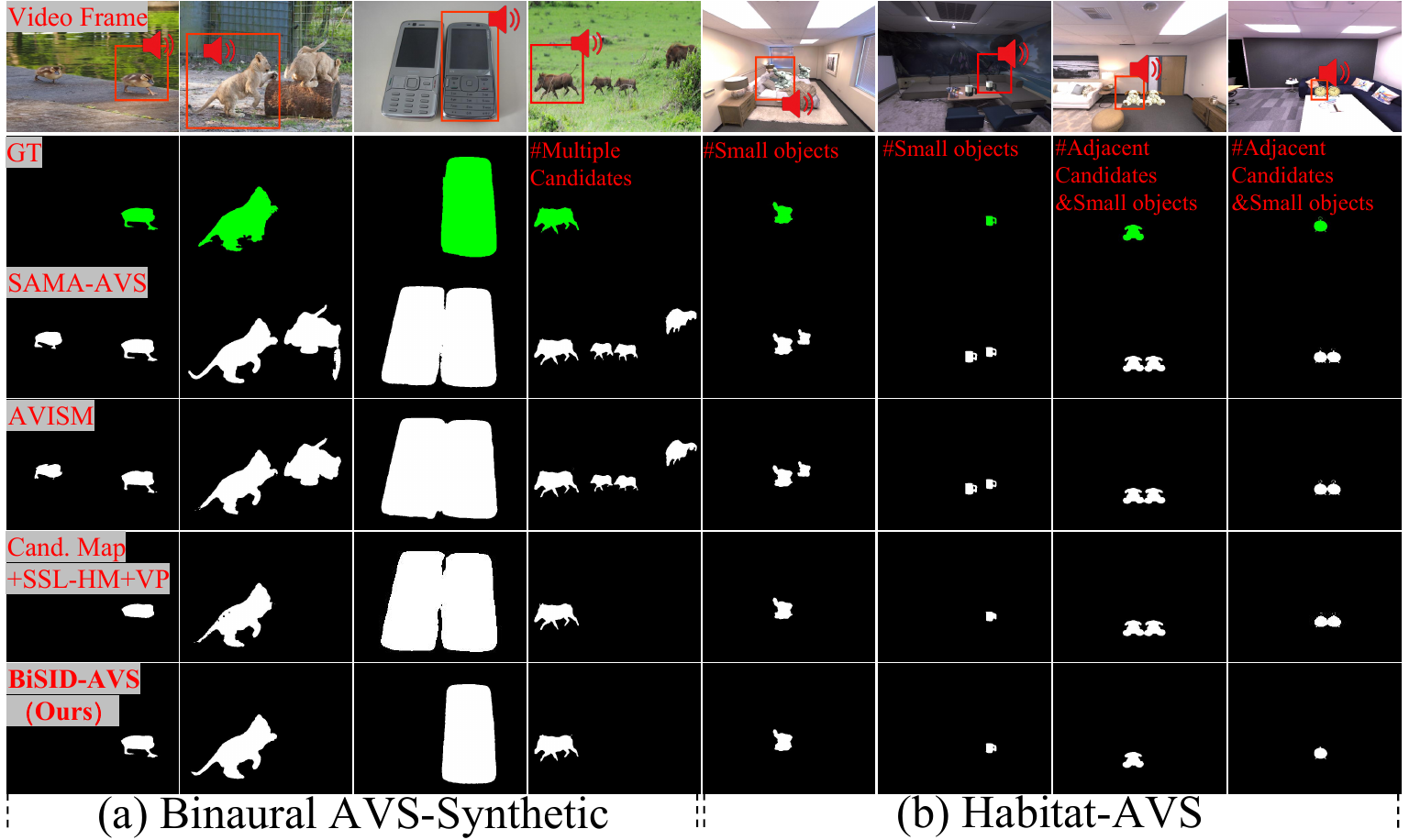}
\caption{Qualitative comparison on representative multi-same-class-candidate scenes on two datasets. }
\label{fig:qualitative_results}
\vspace{-2.5mm}
\end{figure}

\paragraph{Loss function ablation.}
As shown in Table~\ref{tab:ablation_loss}, replacing the bounding-box supervision $\mathcal{L}^{\mathrm{box}}$ with the spatial response map loss $\mathcal{L}^{\mathrm{spl}}$ improves
$\mathrm{mIoU}_{\mathrm{M}}$ and reduces
$F_{\mathrm{same},\mathrm{M}}$, indicating that instance-mask-based spatial supervision is more precise than coarse box supervision. Adding $\mathcal{L}^{\mathrm{sc}}$ further improves both settings by explicitly suppressing same-class distractors. Their combination achieves the best results across all metrics, confirming the complementary effects of spatial-response supervision and same-class suppression.

\begin{table}[t]
\centering
\footnotesize
\setlength{\tabcolsep}{1pt}
\renewcommand{\arraystretch}{1.05}
\begin{tabular*}{\columnwidth}{@{\extracolsep{\fill}}lcccc@{}}
\toprule
Loss setting &
$\mathrm{mIoU}_{\mathrm{M}}^{\uparrow}$ &
$\mathrm{mSILA}_{\mathrm{M}}^{\uparrow}$ &
$F_{\mathrm{same},\mathrm{M}}^{\downarrow}$ &
$\mathrm{SC\mbox{-}DS}_{\mathrm{M}}^{\uparrow}$ \\
\midrule
$\mathcal{L}^{\mathrm{base}}+\mathcal{L}^{\mathrm{box}}$ & 0.8674 & 0.8219 & 0.1047 & 0.8425 \\
$\mathcal{L}^{\mathrm{base}}+\mathcal{L}^{\mathrm{spl}}$ & 0.8791 & 0.8372 & 0.0689 & 0.8679 \\
$\mathcal{L}^{\mathrm{base}}+\mathcal{L}^{\mathrm{box}}+\mathcal{L}^{\mathrm{sc}}$ & \underline{0.8846} & \underline{0.8392} & \underline{0.0428} & \underline{0.8721} \\
$\mathcal{L}^{\mathrm{base}}+\mathcal{L}^{\mathrm{spl}}+\mathcal{L}^{\mathrm{sc}}$ & \textbf{0.8952} & \textbf{0.8564} & \textbf{0.0356} & \textbf{0.8811} \\
\bottomrule
\end{tabular*}
\caption{Loss-function ablation on the Habitat-AVS multi-candidate subset.}
\label{tab:ablation_loss}
\vspace{-1mm}
\end{table}

\subsection{Qualitative Results and Real-Robot Demonstration}
\paragraph{Qualitative comparison.}
Figure~\ref{fig:qualitative_results} shows that SAMA-AVS and AVISM often activate same-class distractors together with the source. Decision-level fusion (Cand. Map + SSL-HM + VP) removes some false regions but remains unreliable for adjacent instances or proposals covering multiple objects. In contrast, BiSID-AVS consistently focuses on the sounding instance and better preserves its contour across both datasets, including small and closely spaced targets. These results are consistent with its lower $F_{\mathrm{same},\mathrm{M}}$ and higher $\mathrm{SC\mbox{-}DS}_{\mathrm{M}}$.

\begin{figure}[t]
\centering
\includegraphics[width=\columnwidth]{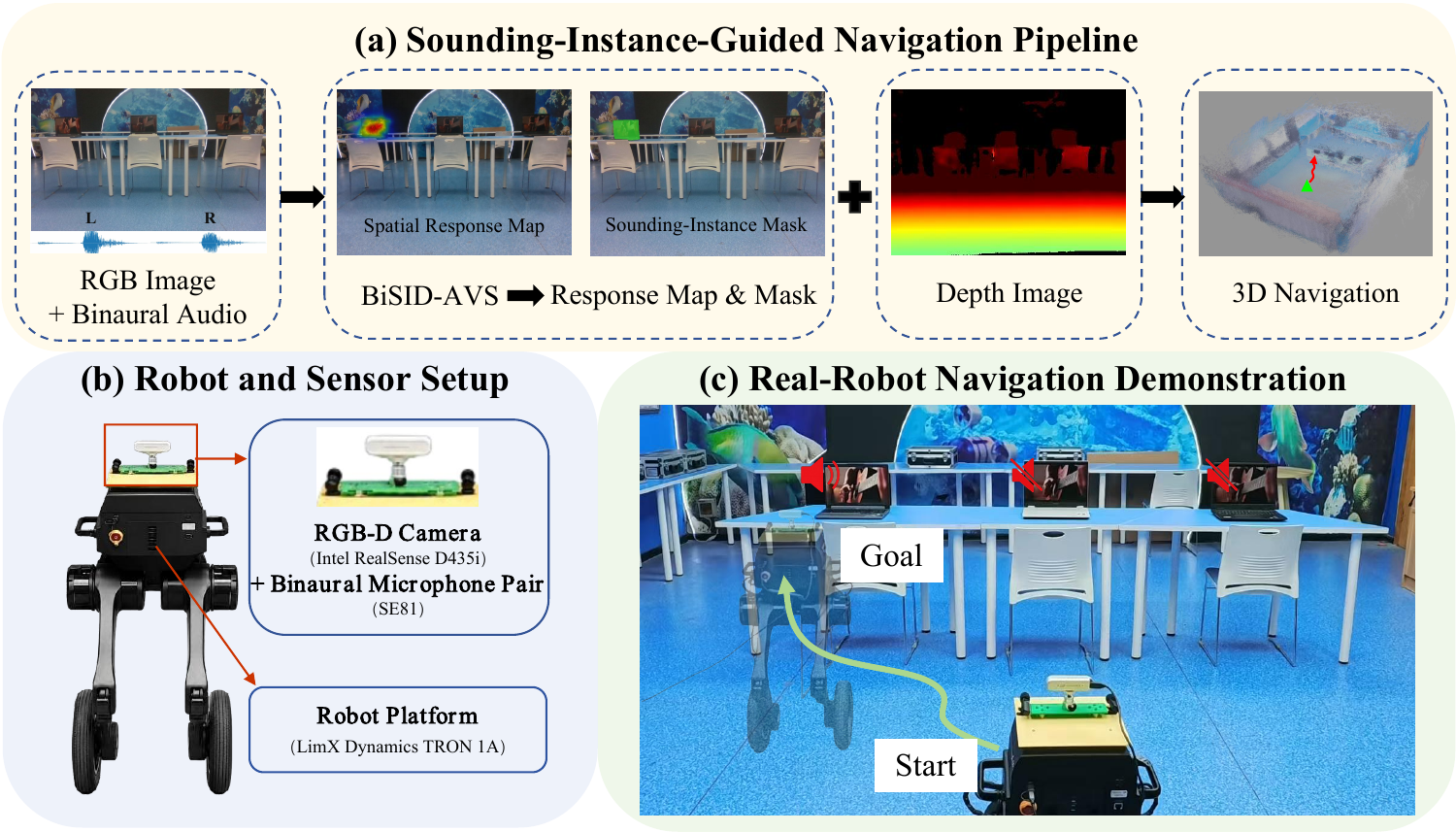}
\caption{Real-robot sounding-instance-guided navigation.}
\label{fig:real_robot_demo}
\vspace{-2mm}
\end{figure}

\paragraph{Real-robot demonstration.}
As shown in Figure~\ref{fig:real_robot_demo}, BiSID-AVS is deployed on a mobile robot equipped with binaural microphones and an RGB-D camera. Among three same-class devices, it identifies the sounding target, whose masked depth is then converted into a navigation goal. This demonstrates the feasibility of using BiSID-AVS as a perceptual front end for robotic navigation toward a sounding object.

\section{Conclusion}
This work establishes binaural sounding-instance disambiguation as a
distinct setting for audio-visual segmentation and introduces BiSID-AVS
to identify the true sounding instance among same-class visual
candidates. By integrating candidate grounding and binaural spatial
response modeling directly into mask prediction, the framework avoids
post-hoc matching between independently estimated source locations and
visual masks. The proposed Binaural AVS-Synthetic and Habitat-AVS
benchmarks, together with instance-level disambiguation metrics, provide
a basis for studying spatially grounded audio-visual segmentation under
controlled and robotic-view conditions. Experiments demonstrate that
BiSID-AVS substantially improves multi-candidate segmentation and
suppresses same-class false activation. The current formulation assumes one active source per audio segment, and future work will focus on multi-source sounding-instance segmentation under overlapping acoustic events.

\begin{small}
\bibliography{references}
\end{small}

\clearpage
\appendix
\setcounter{secnumdepth}{2}
\setcounter{table}{0}
\setcounter{figure}{0}
\setcounter{equation}{0}
\renewcommand{\thetable}{A\arabic{table}}
\renewcommand{\thefigure}{A\arabic{figure}}
\renewcommand{\theequation}{A\arabic{equation}}
\section*{Appendix}

\section{Implementation Details of BiSID-AVS}
\label{sec:supp_implementation}

\subsection{Input Representations}

All RGB images are resized to $512\times512$. The visual backbone is SAM
ViT-H~\cite{kirillov2023sam}, initialized from the official
\texttt{sam\_vit\_h\_4b8939.pth} checkpoint. It uses a patch size of 16, a
token dimension of 1280, 32 Transformer blocks, 16 attention heads, and a
256-channel output neck. The resulting visual feature map has a spatial
resolution of $32\times32$.

The audio-semantic prompting path first averages the synchronized left and
right waveforms. Following the VGGish preprocessing
pipeline~\cite{hershey2017cnn}, the monaural waveform is resampled to 16~kHz
and converted into a $96\times64$ log-Mel patch using a 25-ms analysis window,
a 10-ms hop, and 64 Mel bands spanning 125--7500~Hz. A frozen VGGish encoder
initialized from the \texttt{vggish-10086976.pth} checkpoint pretrained on
AudioSet~\cite{gemmeke2017audioset} produces a 128-dimensional embedding. For
each of the 32 SAM blocks, a
block-specific linear layer maps this embedding from 128 to 512 dimensions,
followed by GELU; a shared linear layer then projects it to the 1280-dimensional
visual-token space. The resulting prompt is added to every visual token before
the corresponding Transformer block. The unprompted visual representation is
computed with the same SAM parameters but without audio-prompt injection.

For binaural spatial modeling, both waveforms are resampled to 16~kHz and
cropped or zero-padded to 2~s. A 66-channel cochleagram spanning 50--8000~Hz
with a sample factor of 1 is computed independently for each ear, and the two
cochleagrams are stacked to form an input of size $2\times66\times T$. The two
ears are never averaged in this path.

\subsection{Spatial Response and Conditioning Modules}

Table~\ref{tab:supp_modules} summarizes the key modules used in BiSID-AVS. The
binaural spatial encoder uses three temporally strided convolutions, each
followed by instance normalization and ReLU. Adaptive average pooling and a
linear projection produce a 256-dimensional spatial audio query. A channel-wise
inner product between this query and the stop-gradient unprompted visual feature
plus SAM positional encoding gives the $32\times32$ source response. The dense
prompt mapper converts the one-channel response into a 256-channel prompt for
the SAM mask decoder.

\begin{table*}[!t]
\centering
\small
\setlength{\tabcolsep}{5pt}
\renewcommand{\arraystretch}{1.12}
\begin{tabular}{@{}
>{\raggedright\arraybackslash}m{0.19\textwidth}
>{\centering\arraybackslash}m{0.14\textwidth}
>{\centering\arraybackslash}m{0.55\textwidth}
@{}}
\toprule
Module & Dimensions & Configuration \\
\midrule
Spatial convolution 1
& $2\rightarrow32$
& Kernel $1\times31$; stride $1\times10$; padding $0\times15$; IN; ReLU \\
Spatial convolution 2
& $32\rightarrow64$
& Kernel $3\times11$; stride $1\times5$; padding $1\times5$; IN; ReLU \\
Spatial convolution 3
& $64\rightarrow128$
& Kernel $3\times7$; stride $2\times4$; padding $1\times3$; IN; ReLU \\
Spatial query projection
& $128\rightarrow256$
& Adaptive $1\times1$ pooling followed by a linear projection \\
Dense prompt mapper
& $1\rightarrow64\rightarrow256$
& Two $3\times3$ convolutions with GELU between them \\
Candidate-region auxiliary head
& $256\rightarrow64\rightarrow1$
& $3\times3$ convolution, ReLU, and $1\times1$ convolution \\
SAM mask decoder
& 256-D inputs
& Two-way Transformer: depth 2, 8 heads, MLP dimension 2048 \\
\bottomrule
\end{tabular}
\caption{Implementation specifications of the key BiSID-AVS modules.}
\label{tab:supp_modules}
\end{table*}

Feature-level conditioning converts the spatial response into a bounded sigmoid
gate and applies it to the detached audio-prompted candidate feature through
$1+\alpha G^{\mathrm{spl}}$. The scalar $\alpha$ is a single globally shared,
unconstrained learnable parameter initialized to zero, so the module begins as
an identity mapping. Decoder-level conditioning uses the mapped spatial response
as the dense prompt. The sparse-prompt input is an empty tensor; no point or box
prompt is used.

\subsection{Optimization and Inference}

VGGish and the non-prompt parameters of the SAM visual encoder are frozen.
Training updates the audio-prompt mappings, spatial encoder and query
projection, candidate-region auxiliary head, dense prompt mapper, spatial gate, and
SAM mask decoder. We train for 40 epochs with AdamW, an initial learning rate
of $2\times10^{-4}$, weight decay of 0.01, cosine decay to $10^{-7}$ without
warm-up, batch size 1, a fixed random seed of 42, and one NVIDIA RTX 4090 GPU.
The loss weights are
$\lambda^{\mathrm{ins}}=1.0$, $\lambda^{\mathrm{cand}}=1.0$,
$\lambda^{\mathrm{spl}}=0.5$, $\lambda^{\mathrm{sc}}=0.1$, and
$\beta^{\mathrm{non}}=1.0$. Model selection uses the validation partition of
each fold, with $\mathrm{mIoU}_{\mathrm{All}}$ as the selection metric. At
inference, the instance-mask logits are converted to probabilities with a
sigmoid and binarized at 0.5 for evaluation.

\subsection{Loss and Ablation Details}

The soft IoU loss used for the sounding-instance and candidate-region
predictions is one minus the standard soft intersection-over-union and is
averaged over the batch. For the candidate-region BCE, the positive-class
weight is $\min(N_{-}/(N_{+}+\epsilon),100)$, where $N_{+}$ and $N_{-}$
denote the foreground and background pixel counts and $\epsilon$ prevents
division by zero. The spatial response loss separately normalizes its source
and non-source terms, as defined in the main paper.

For the loss ablation, $\mathcal{L}^{\mathrm{base}}
=\lambda^{\mathrm{ins}}\mathcal{L}^{\mathrm{ins}}
+\lambda^{\mathrm{cand}}\mathcal{L}^{\mathrm{cand}}$ collects the
instance-mask and candidate-region objectives.
$\mathcal{L}^{\mathrm{box}}$ denotes the alternative bounding-box spatial
supervision: a binary target is set to one inside the sounding-instance box
and zero elsewhere, and binary cross-entropy is applied to the spatial-response
logits. The fixed loss weights are omitted from the ablation-table notation
for clarity.

\section{Construction of Binaural AVS-Synthetic}
\label{sec:supp_synthetic}

\subsection{Pseudo-Instance Annotation}

Binaural AVS-Synthetic is derived from AVS-Synthetic as used by
SAMA-AVS~\cite{liu2024sama}. The original semantic mask covers all visible
regions of the annotated sound category. We apply 8-connected component
analysis to this mask and treat the valid connected components as same-class
instance candidates. A component is retained only if its area is at least 100
pixels and at least 0.5\% of the total semantic foreground. Samples with no
valid component or more than five valid components are removed. A component is
then sampled with random seed 42 as the sounding instance. The selected
component must contain at least 100 pixels and have a bounding-box width and
height of at least five pixels. The selected mask is resized to the RGB-image
resolution with nearest-neighbor interpolation; its box and centroid are
recorded as the sounding-instance annotation.

Connected components are only an initial proposal mechanism, so we perform an
additional category-consistency check with GroundingDINO
~\cite{liu2024groundingdino}. The normalized category name is used as the text
query, with both box and text thresholds set to 0.25. We retain a sample only
when the number of GroundingDINO detections equals the number of valid mask
components. This filtering reduces the 55,705 samples that pass component-level
quality control to 31,045 verified samples. The final set contains 4,488
multi-same-class-candidate samples (14.5\%).

\subsection{Spatially Consistent Binaural Rendering}

The dry audio is bound by the pair of image identifier and sound category,
which prevents an image containing multiple annotated categories from receiving
audio from the wrong category. Depth Anything~\cite{yang2024depthanything}
estimates a relative depth map for each image. We normalize the depth map within
each image and sample the value at the selected instance centroid.

Let $(c_x,c_y)$ be the selected centroid in an image of width $W$ and height
$H$. Its normalized image coordinates are
\begin{equation}
\begin{aligned}
x_n&=\operatorname{clip}\!\left(\frac{c_x-W/2}{W/2},-1,1\right),\\
y_n&=\operatorname{clip}\!\left(\frac{H/2-c_y}{H/2},-1,1\right).
\end{aligned}
\end{equation}
We map these coordinates to azimuth and elevation as
\begin{equation}
\theta=70^{\circ}x_n,\qquad \phi=40^{\circ}y_n.
\end{equation}
The normalized depth is mapped to a nominal distance in $[0.8,2.2]$~m. We use
the nearest measured direction in the SADIE D1 HRTF
database~\cite{armstrong2018sadie,sadie2024database} and convolve the dry
waveform with the corresponding left and right head-related impulse responses.
Each binaural waveform is cropped or zero-padded to 2~s and peak-normalized to
0.95. Finally, the same 16-kHz, 66-band cochleagram extraction described in
the implementation section is applied to both channels.

\subsection{Annotations and Cross-Validation}

Each sample contains an RGB image, synchronized binaural waveform, stacked
cochleagram, sounding-instance mask and box, and a same-class candidate mask.
The candidate mask is used only as auxiliary supervision and for computing the
same-class evaluation metrics. The five-fold protocol groups records originating
from the same original sample and stratifies the groups by sound category and
single-/multi-candidate status. For every fold, 10\% of the remaining training
groups are reserved for validation; each metric is computed separately within each held-out fold on its
applicable evaluation subset (single-candidate, multi-candidate, or full set).
The reported result is the unweighted mean of the five fold-specific scores.

\section{Construction of Habitat-AVS}
\label{sec:supp_habitat}

\subsection{Scenes, Viewpoints, and Assets}

Habitat-AVS is generated with Habitat-Sim, Replica, and the SoundSpaces audio
sensor~\cite{savva2019habitat,straub2019replica,chen2020soundspaces,chen2022soundspaces2}.
It contains four Replica scenes (\textsf{room\_0}, \textsf{room\_1},
\textsf{office\_0}, and \textsf{office\_2}), ten predefined robotic
viewpoints, and 14 object categories. The viewpoint identifiers are
\textsf{room0\_0--2},
\textsf{room1\_0--1}, \textsf{office0\_0--1}, and
\textsf{office2\_0--2}. Each viewpoint specifies a camera position, a look-at
point, and a three-dimensional placement range inside the camera field of
view. Camera positions are perturbed by at most 0.2~m horizontally and
0.05~m vertically to increase viewpoint variation while preserving the
viewpoint identity used for data splitting.

Single-candidate samples contain one target-class object. Multi-candidate
samples contain two objects of the same class, of which the first is assigned
as the sounding source and the second remains silent. The objects are placed
on surfaces found by downward ray casting. Five support rays---one at the
center and four around the footprint---must all intersect the scene, and the
maximum difference among their hit heights must not exceed 0.05~m.

\subsection{Visibility Control and Ground Truth}

Every generated observation passes the following deterministic checks before
being saved:
\begin{itemize}
\item each target instance contains at least 100 visible pixels;
\item its visible mask remains more than five pixels away from every image
boundary;
\item the visible-mask fill ratio within its 2D bounding box is at least 0.40;
and
\item the 2D bounding boxes of all inserted objects do not overlap.
\end{itemize}
If any check fails, object placement is resampled, with at most 50 attempts per
sample. Habitat's semantic identifiers directly provide the pixel-level
ground truth. The sounding-instance mask contains only the source identifier,
whereas the same-class candidate mask is the union of all target-class
identifiers. The sounding-instance bounding box is computed from its visible
pixels.

\subsection{Binaural Audio and Evaluation Splits}

The audio source is placed at the three-dimensional location of the sounding
object. SoundSpaces renders a two-channel room impulse response at 44.1~kHz.
For this benchmark, indirect reflections and material-dependent acoustics are
disabled so that the generated cue is dominated by source--receiver geometry.
The category-matched dry waveform is convolved separately with the left and
right impulse responses and peak-normalized. The rendering routine writes a
1-s binaural clip; before feature extraction, it is zero-padded to 2~s and
processed into the semantic and spatial audio inputs described above.

The resulting dataset contains 11,781 samples from 14 categories; 45.3\% are
multi-candidate samples. To prevent nearly identical observations from the
same robotic viewpoint from appearing in both training and test data, the ten
viewpoints are organized into five fixed pairs. Each fold holds out one pair
for testing, and 10\% of the remaining viewpoint-grouped data is used for
validation. For each fold, a fold-specific annotation file assigns every sample to the
training, validation, or test split. As with Binaural AVS-Synthetic, each metric is computed separately within
each held-out fold on its applicable evaluation subset, and the reported
result is the unweighted mean of the five fold-specific scores.

For scene-disjoint evaluation, we additionally construct four
leave-one-scene-out folds using \textsf{room\_0}, \textsf{room\_1},
\textsf{office\_0}, and \textsf{office\_2}. In each fold, all viewpoints and
samples from one scene are held out for testing, while the other three scenes
form the training pool. Ten percent of this pool is reserved for validation,
and the checkpoint with the best validation
$\mathrm{mIoU}_{\mathrm{All}}$ is evaluated on the held-out scene. Thus, no
viewpoint from the test scene appears in training or validation.

\section{Baseline Adaptation and Training Protocols}
\label{sec:supp_baselines}

\subsection{Unified Task Adaptation}

TPAVI~\cite{zhou2022avs}, AVSegFormer~\cite{gao2023avsegformer},
COMBO~\cite{yang2024combo}, VCT~\cite{huang2025vct},
SAMA-AVS~\cite{liu2024sama}, and AVISM~\cite{guo2025avism} are retrained
separately on Binaural AVS-Synthetic and Habitat-AVS using the same five-fold
partitions as BiSID-AVS. Their final prediction target is always the true
sounding-instance mask, rather than the candidate mask covering all same-class
candidates.
Monaural methods receive the channel average and retain their official audio
preprocessing. Methods requiring a video receive a static clip formed by
repeating the same RGB image to the clip length expected by their public
implementation. The same instance mask is copied to every frame, and no
frame-dependent crop, geometric transform, or appearance perturbation is
applied. Consequently, the adaptation satisfies the original interfaces
without adding motion or cross-frame visual evidence.

We retain the public architecture and recommended optimization strategy of
each baseline, except for the input/output changes required by static RGB,
binaural-to-monaural conversion, and instance-mask supervision. Model
selection is based only on the validation partition of the corresponding
fold. All methods use the same sigmoid threshold of 0.5 and are evaluated by
the mIoU, mSILA, $F_{\mathrm{same}}$, and SC-DS implementations used for
BiSID-AVS.

\subsection{AVISM Adaptation}

AVISM~\cite{guo2025avism} is designed for video-level audio-visual instance
segmentation. For each image, we construct a static clip by repeating it to
the native clip length $T$ of the public AVISM configuration. Its
audio input is the binaural-channel average, and the sounding-instance mask is
repeated over all $T$ frames during training. Following the AVISM evaluation
interface, the true sound category and the number of sounding instances
$K=1$ are supplied, but no true mask, target box, or target identity is
provided at inference. We select the predicted trajectory with the highest
sounding confidence and evaluate its mask on the reference frame. This result
therefore measures AVISM under the single-observation setting studied in the
paper; it should not be interpreted as an upper bound on AVISM in its native
dynamic-video setting.

\subsection{Decision-Level Fusion Baselines}

The decision-level baselines use the frozen SAM ViT-H automatic mask generator
with its default configuration to obtain category-agnostic visual proposals.
The BiSID-AVS spatial response map is upsampled to the image resolution and
treated as a sound source localization heatmap (SSL-HM). For a proposal mask
$V_j$ and response map $R$, the proposal score is
\begin{equation}
s_j(R)=
\frac{\sum_{(u,v)}V_j(u,v)R(u,v)}
{\sum_{(u,v)}V_j(u,v)+\epsilon}.
\end{equation}
Here, $\epsilon$ prevents division by zero.
SSL-HM + VP uses $R=\hat{S}^{\mathrm{spl}}$ and outputs the proposal with the
largest score. Cand. Map + SSL-HM + VP first weights the response by the
predicted candidate-region probability map,
$R=\hat{S}^{\mathrm{spl}}\odot\hat{M}^{\mathrm{cand}}$, before applying the
same proposal-selection rule. The oracle variant replaces the predicted
candidate map with the ground-truth candidate mask,
$R=\hat{S}^{\mathrm{spl}}\odot M_{\mathrm{gt}}^{\mathrm{cand}}$.
All three variants therefore introduce spatial evidence only at the output
stage and do not condition visual feature formation or mask decoding.

\subsection{Interaural Counterfactual Evaluation}

The interaural counterfactual evaluation is performed on the Habitat-AVS
multi-candidate test samples without retraining the model. For the
channel-swapping condition, the left and right waveforms of each sample are
exchanged while its RGB image remains unchanged. For the mismatched-audio
condition, each test image is paired with one binaural waveform randomly drawn
from another sample of the same scene and sound category but a different source
location. Eligible donors are sampled once using random seed 42, yielding one
fixed mismatched waveform per test sample. This construction preserves
sound-category information while disrupting the correspondence between the
binaural spatial cues and the visual source location.

\end{document}